\documentclass[10pt,a4paper]{article}
\usepackage{PRIMEarxiv}
\usepackage[utf8]{inputenc} 
\usepackage[T1]{fontenc}    
\usepackage{hyperref}       
\usepackage{url}            
\usepackage{booktabs}       
\usepackage{amsfonts}       
\usepackage{nicefrac}       
\usepackage{microtype}      
\usepackage{lipsum}
\usepackage{float}
\usepackage{fancyhdr}       
\usepackage{graphicx}       
\graphicspath{{media/}}     

\usepackage{amsmath}
\usepackage[numbers,sort&compress]{natbib}
\usepackage{xurl}   
\usepackage{rotating} 
\usepackage{listings} 
\usepackage{xcolor}   

\usepackage[utf8]{inputenc}
\usepackage{amsmath, amsthm, amssymb}
\usepackage{graphicx}
\usepackage{booktabs} 

\usepackage[nottoc,notlot,notlof]{tocbibind} 
\usepackage{caption} 
\usepackage{tikz}
\usepackage{amsmath}
\usetikzlibrary{shapes.geometric, arrows.meta, positioning, fit}

\usepackage{tikz}
\usepackage{float} 

\usetikzlibrary{positioning, shapes.geometric, arrows.meta, fit}

\tikzstyle{startstop} = [rectangle,  rounded corners=4pt, minimum width=2.5cm, minimum height=0.8cm, text centered, draw=black, fill=gray!20]
\tikzstyle{process} = [rectangle,  rounded corners=4pt, minimum width=3cm, minimum height=1cm, text centered, draw=black, fill=blue!20, text width=4.5cm, align=center]
\tikzstyle{sigmoid} = [rectangle,  rounded corners=4pt,  minimum width=1.5cm, minimum height=0.8cm, text centered, draw=black, fill=green!20, text width=1.8cm, align=center]
\tikzstyle{labelbox} = [rectangle,  rounded corners=4pt, minimum width=1.5cm, minimum height=0.8cm, text centered, draw=black, fill=orange!20, text width=2cm, align=center]
\tikzstyle{arrow} = [thick, -{Stealth[length=2mm, width=1.5mm]}] 

\hypersetup{
    colorlinks=true,
    linkcolor=blue,
    filecolor=magenta,      
    urlcolor=cyan,
    citecolor=red,
}

\definecolor{codegreen}{rgb}{0,0.6,0}
\definecolor{codegray}{rgb}{0.5,0.5,0.5}
\definecolor{codepurple}{rgb}{0.58,0,0.82}
\definecolor{backcolour}{rgb}{0.95,0.95,0.92}
\lstdefinestyle{mystyle}{
    backgroundcolor=\color{backcolour},   
    commentstyle=\color{codegreen},
    keywordstyle=\color{magenta},
    numberstyle=\tiny\color{codegray},
    stringstyle=\color{codepurple},
    basicstyle=\ttfamily\footnotesize,
    breakatwhitespace=false,         
    breaklines=true,                 
    captionpos=b,                    
    keepspaces=true,                 
    numbers=left,                    
    numbersep=5pt,                  
    showspaces=false,                
    showstringspaces=false,
    showtabs=false,                  
    tabsize=2
}
\title{\textbf{Paladin-mini: A Compact and Efficient Grounding Model Excelling in Real-World Scenarios}}

\author{
  Dror Ivry \\
   \And
  Oran Nahum \\
}

\date{\vspace{-5ex}} 

\begin{document}

\maketitle

\begin{abstract}
This paper introduces two significant contributions to address the issue of grounding claims in a given context. Grounding means that given a document D and a claim c, there’s at least one supportive evidence for the claim in the document. We will introduce Paladin-mini, a compact (3.8B parameters) open-source classifier model (used for labeling data as grounded or ungrounded) engineered for robust performance in real-world scenarios, and the grounding-benchmark, a new evaluation dataset designed to assess performance on critical reasoning tasks. We’ll also demonstrate the results of Paladin-mini with benchmarks against the current State-of-the-art and share clear and reproducible results.
\end{abstract}

\section{Introduction}
The widespread integration of Large Language Models (LLMs) into high-stakes professional applications is critically hampered by their tendency to produce non-factual or "ungrounded" outputs. To address this challenge, we introduce PALADIN, a suite of models specifically engineered for grounding—the process of verifying a model's claims against a given set of source documents. This paper presents Paladin-mini, a compact, efficient, and open-source grounding model built upon Microsoft's Phi-4-mini-instruct. Through a specialized training regimen incorporating both public and synthetic datasets, Paladin-mini is optimized for robust performance in practical scenarios. To rigorously evaluate these capabilities, we also introduce the grounding-benchmark, a novel benchmark focused on critical real-world tasks such as numerical, temporal, and closed-domain reasoning. Our findings demonstrate that Paladin-mini significantly outperforms larger, state-of-the-art models on this specialized benchmark, representing a key advancement in the development of reliable and trustworthy AI for widespread enterprise deployment.

\section{Related Work}
\subsection{Grounding and Fact-Checking Models}
A prominent recent development is the MiniCheck family of models, introduced by Tang et al. (2024). MiniCheck focuses on leveraging synthetic data to train compact models for fact verification. Key to its methodology are two synthetic data generation techniques: Claim to Doc (C2D), which generates a supporting document for a given claim, and Doc to Claim (D2C), which generates a claim from a document. These techniques enable the training of smaller models to verify multiple facts within a sentence against multiple sentences in grounding documents. The MiniCheck suite includes variants such as MiniCheck-Flan-T5-Large (770M parameters) and the more powerful Bespoke-MiniCheck-7B (7B parameters). These models have demonstrated strong performance on the LLM-AggreFact benchmark, with some achieving accuracy comparable to that of GPT-4.

While the generalist approach of MiniCheck, using broad synthetic data generation, has proven effective for improving performance on aggregated benchmarks, it lacks the nuances needed for real world use cases. Here is some anecdotal evidence of that behavior - note that we’ll test that more empirically in the later sections.

\begin{figure}[h!]
\centering
\begin{verbatim}
prompt = '''A toy store sells LEGO sets for $80 and 
remote control cars for $120. Buying 3 or more 
items gets a 15% discount.'''

response = "Buying 2 LEGO sets and 1 remote control 
car costs $238 after the 15% discount."

minicheck pred: UNGROUNDED
\end{verbatim}
\caption{An anecdotal example of MiniCheck's performance on a quantitative reasoning task. MiniCheck incorrectly labels a mathematically correct calculation as 'UNGROUNDED,' highlighting a potential weakness in nuanced, multi-step reasoning that the PALADIN framework aims to address.}
\label{fig:minicheck_example}
\end{figure}

The work presented in this paper builds upon the idea of using synthetic data but takes a more targeted approach. Instead of generating data for general-purpose fact-checking, the PALADIN framework utilizes synthetic datasets specifically engineered to address the types of errors—numerical, logical, and temporal—that are under-represented in general benchmarks but are critical in real-world applications. This distinction between generalist and specialist synthetic data is a core differentiator, positing that for high-stakes, practical applications, specialized training is necessary to achieve the required level of reliability.

\subsection{Benchmarking Factual Consistency}
The evaluation of grounding models relies heavily on robust benchmarks. The LLM-AggreFact benchmark has emerged as a significant resource, unifying 11 publicly available datasets to create a comprehensive testbed for fact verification. Other notable benchmarks include TofuEval, which focuses on hallucinations in topic-focused dialogue summarization; ClaimVerify, which assesses verifiability in the outputs of generative search engines; and FactCheck-GPT, designed for document-level fact-checking.

Despite their utility, these existing benchmarks primarily provide an aggregated measure of performance. Their broad-stroke evaluations can mask critical weaknesses in specific reasoning domains. For instance, a model might achieve a high overall score on LLM-AggreFact while being unable to perform simple arithmetic or correctly compare two dates. This limitation highlights the need for more specialized, application-oriented evaluation methodologies. The evolution towards more focused benchmarks, such as the grounding-benchmark introduced in this work, reflects a maturing understanding of LLM evaluation needs.

\section{The PALADIN Grounding Framework}
\subsection{Model Architectures and Training}
The PALADIN suite comprises two models engineered with different optimization goals: Paladin-mini, designed for efficiency and open-source accessibility, and Paladin-large, a private model optimized for state-of-the-art accuracy (we’ll address that model in a future paper).

Paladin-mini (Open Source) is built upon microsoft/Phi-4-mini-instruct, a 3.8B parameter model known for its strong reasoning capabilities and a 128K token context length. It is designed for environments with memory and compute constraints. The model undergoes full Supervised Fine-Tuning (SFT) on a curated dataset. It is not quantized during training and uses float16 precision for inference, resulting in a memory footprint of approximately 7GB. This approach prioritizes maintaining the model's generative and instruction-following capabilities alongside its grounding function. The distinct specifications of the two models are summarized in Table 1.

\begin{table}[h!]
\centering
\caption{Paladin Model Specifications. This table provides a side-by-side comparison of the two models, highlighting their different design philosophies and technical specifications.}
\begin{tabular}{@{}ll@{}}
\toprule
\textbf{Feature} & \textbf{Paladin-mini (Open Source)} \\ \midrule
Base Model & microsoft/Phi-4-mini-instruct \\
Training Type & Full training (SFT) \\
Task Type & Supervised Fine-Tuning \\
Training Dataset Size & "23,000 samples" \\
Quantized During Training & No \\
Inference Size & "~7GB [float16]" \\ \bottomrule
\end{tabular}
\end{table}

\subsection{Specialized Training Corpus}
Paladin-mini is trained on a curated dataset of 23,000 samples. This training corpus is a strategic combination of established, publicly available academic datasets and synthetic data. The public portion includes data from sources like MiniCheck, and AggreFact, ensuring the models develop a strong baseline in general fact verification. The crucial component, however, is the data we synthesized. This synthetic data is not general-purpose; it is specifically engineered to target the types of errors most critical in real-world applications. The dataset includes samples from proprietary collections such as real\_world\_use\_cases\_ungrounded\_samples, df\_time\_and\_dates\_fix\_errors\_df\_ungrounded, and df\_prices\_fix\_errors\_df\_ungrounded. The creation of such targeted synthetic data, often using powerful LLMs as generators, is a modern and effective technique for enhancing model robustness in specific domains. This targeted data is the primary driver of Paladin-mini's enhanced performance on the specialized categories within the grounding-benchmark, differentiating it from models trained primarily on general fact-checking corpora.

\subsection{A Formalism for Synthetic Data Generation}
To ensure the logical consistency and quality of the proprietary synthetic data, a formal methodology was created. The original formulation presented in the source document contained notational and logical errors. What follows is a corrected, mathematically rigorous framework for defining groundedness and generating synthetic examples, based on standard principles of formal methods and logic. This formalism provides the theoretical backbone for the data generation process, ensuring that the models are trained on logically consistent and complex examples.

First, we define the core components:
\begin{itemize}
    \item \textbf{Evidence Set (D):} Let $D = \{s_1, s_2, \ldots, s_m\}$ be a finite set of evidence sentences extracted from a source document.
    \item \textbf{Claim (c):} Let a claim $c$ be a logical proposition. For complex claims, it is useful to decompose them into a conjunction of atomic facts, such that $c \leftrightarrow (a_1 \land a_2 \land \ldots \land a_n)$. Each atomic fact represents a single, verifiable statement.
\end{itemize}

With these definitions, we can formally define the concept of grounding:
\begin{itemize}
    \item \textbf{Grounding:} A claim $c$ is considered \textbf{grounded} in an evidence set D if and only if D logically entails c. This relationship is denoted using the semantic entailment symbol: $D \models c$.
    \item Conversely, a claim is \textbf{ungrounded} if the evidence set does not entail it, denoted as $D \not\models c$. This means that in every possible interpretation where all sentences in D are true, c must also be true for the claim to be grounded.
\end{itemize}

To generate high-quality and challenging training examples, it is crucial to ensure that the supporting evidence is both necessary and non-redundant. This is achieved through a minimality condition:
\begin{itemize}
    \item \textbf{Minimal Support:} For a complex claim $c \leftrightarrow (a_1 \land \ldots \land a_n)$, let each atomic fact $a_i$ be supported by a minimal subset of evidence $D_i \subseteq D$. The evidence set D provides minimal support for the claim c if, for every atomic fact $a_i$, the full claim c is not entailed when the specific evidence for $a_i$ is removed. This can be expressed formally as: $D' = \forall i \in \{1, \ldots, n\} : (D \setminus D_i) \not\models c$. This condition ensures that every piece of supporting evidence is essential for verifying the full claim.
\end{itemize}

Moreover by doing so we can ensure that c is UNGROUNDED in D’. By adhering to this formal framework, the synthetic data generation process can produce logically sound triplets of claims and documents and a label (D’, c, label) we can then use those triplets to train our model using standard SFT practices and produce a model capable of classifying claims as grounded or ungrounded in a given document.

\section{Benchmarking for Real-World Grounding}
\subsection{The Grounding-Benchmark}
To address the limitations of existing evaluation methods, this work introduces the Qualifire-grounding-benchmark, publicly available on Hugging Face. The benchmark's primary objective is to assess a model's ability to perform grounding on tasks that are critical in practical, high-stakes enterprise applications, thereby moving beyond a single aggregated score to a more nuanced profile of a model's capabilities. It comprises four distinct evaluation categories, each designed to probe a specific reasoning skill:
\begin{itemize}
    \item \textbf{General:} This category tests a model's ability to comprehend and apply abstract logical structures, such as entailment and contradiction, to novel topics. It measures true generalization by evaluating whether a model can recognize a reasoning pattern independent of the specific content. For example, it might test if a model can identify that a claim about "sustainable packaging" is directly supported by a sentence stating the same fact using different words.
    \item \textbf{Logical:} This category assesses fine-grained reasoning within specific, often technical or industry-related contexts. It measures the model's ability to perform precise fact-checking against documents containing specialized information, a crucial capability for reliable deployment in enterprise environments. An example could involve verifying a claim about quantum-resistant cryptography against a technical description from NIST.
    \item \textbf{Prices \& Math:} This category evaluates quantitative reasoning, a non-negotiable skill for financial, e-commerce, and other data-driven applications. It measures the model's precision in performing multi-step arithmetic calculations based on data such as rates, taxes, and fees to confirm the accuracy of a financial claim.
    \item \textbf{Time \& Dates:} This category benchmarks temporal reasoning, a critical skill for any application involving scheduling, historical analysis, or sequential data. It tests the model's ability to accurately parse, compare, and order dates and times from a text to verify claims, ensuring reliability in time-sensitive tasks.
\end{itemize}

\subsection{Experimental Setup}
The experimental evaluation was conducted using a suite of models, including Paladin-mini, its private counterpart Paladin-large, and the primary competitor, Bespoke-MiniCheck-7B. To provide broader context, the Gemini-2.0-flash model was also included in the comparison. The models were evaluated on two main benchmarks: the newly proposed Qualifire-grounding-benchmark and a subset of eight datasets from the established LLM-AggreFact benchmark. The use of LLM-AggreFact serves to measure general fact-checking competence and provides a point of comparison with the wider academic literature.

The primary evaluation metric used across all experiments is \textbf{Balanced Accuracy (BACC)}. This metric is defined as the arithmetic mean of sensitivity (True Positive Rate or Recall) and specificity (True Negative Rate):

$$ BACC = \frac{\text{Sensitivity} + \text{Specificity}}{2} = \frac{\frac{TP}{TP+FN} + \frac{TN}{TN+FP}}{2} $$

In classification tasks like fact-checking, the distribution of "grounded" and "ungrounded" labels can be imbalanced. Standard accuracy can be misleading in such cases, as a model that simply predicts the majority class can achieve a high score while being practically useless. BACC mitigates this by giving equal weight to the model's performance on each class, providing a more reliable and interpretable measure of performance, especially when the cost of misclassifying the minority class is high.

\section{Results and Analysis}
The empirical results from the comparative evaluation are presented in Table 2 and Table 3. The analysis focuses on the relative performance of Paladin-mini and its primary competitor, Bespoke-MiniCheck-7B, across both the specialized Qualifire benchmark and the general LLM-AggreFact subsets.

\begin{table}[h!]
\centering
\caption{Comparative Performance on Qualifire Benchmark Categories (BACC \%). The average (AVG.) column represents the mean BACC across all benchmarks in both Table 2 and Table 3. Paladin-mini and its main competitor are highlighted.}
\begin{tabular}{@{}lcccccc@{}}
\toprule
\textbf{Model} & \textbf{General} & \textbf{Logical} & \textbf{Time/Dates} & \textbf{Prices/Math} & \textbf{AVG.} & \textbf{Params} \\ \midrule
paladin-large (private) & 91.97 & 98.2 & 91.0 & 98.0 & 94.792 & 14B \\
\textbf{paladin-mini} & \textbf{91.97} & \textbf{97.1} & \textbf{82.0} & \textbf{96.0} & \textbf{91.767} & \textbf{3.8B} \\
\textbf{Bespoke-MiniCheck-7B} & \textbf{84.02} & \textbf{92.8} & \textbf{90.0} & \textbf{46.0} & \textbf{78.205} & \textbf{7B} \\ \bottomrule
\end{tabular}
\end{table}

\subsection{Analysis of Qualifire Benchmark Performance}
The results on the Qualifire-grounding-benchmark, shown in Table 2, reveal significant distinctions in model capabilities. The comparison between Paladin-mini and the larger Bespoke-MiniCheck-7B is particularly illuminating.
\begin{itemize}
    \item \textbf{Areas of Strength:} Paladin-mini demonstrates substantially superior performance in three of the four categories. In General reasoning, it achieves a BACC of 91.97\%, nearly 8 points higher than Bespoke-MiniCheck-7B's 84.02\%. In the Logical category, it scores 97.1\% compared to 92.8\%. The most dramatic difference is observed in the Prices \& Math category, where Paladin-mini attains an exceptional BACC of 96.0\%, while Bespoke-MiniCheck-7B scores only 46.0\%, a performance level barely above random chance. This stark contrast directly validates the hypothesis that specialized training data targeting numerical reasoning leads to specialized, high-performance capabilities.
    \item \textbf{Area for Improvement:} In the Time \& Dates category, Bespoke-MiniCheck-7B outperforms Paladin-mini, scoring 90.0\% to Paladin-mini's 82.0\%. This indicates that the proprietary training data for temporal reasoning may be less effective than for other categories, or that the competitor model has an inherent architectural advantage in this specific domain. This is candidly identified as an area for future work.
\end{itemize}

\begin{table}[h!]
\centering
\caption{Comparative Performance on LLM-AggreFact Subsets (BACC \%). Paladin-mini and its main competitor are highlighted.}
\begin{tabular}{@{}lccccccccc@{}}
\toprule
\textbf{Model} & \textbf{AVG} & \textbf{AF CNN} & \textbf{AF XSum} & \textbf{Tofu M.} & \textbf{Tofu B.} & \textbf{Wice} & \textbf{Reveal} & \textbf{Claim V.} & \textbf{FactCk.} \\ \midrule
paladin-large (private) & 77.825 & 64.01 & 74.77 & 74.76 & 79.56 & 78.63 & 90.77 & 80.14 & 79.96 \\
\textbf{paladin-mini} & \textbf{73.083} & \textbf{59.81} & \textbf{71.05} & \textbf{69.25} & \textbf{71.91} & \textbf{71.63} & \textbf{89.44} & \textbf{75.32} & \textbf{76.26} \\
\textbf{Bespoke-MiniCheck-7B} & \textbf{77.7} & \textbf{65.50} & \textbf{77.80} & \textbf{76.00} & \textbf{78.30} & \textbf{83.00} & \textbf{88.00} & \textbf{75.30} & \textbf{77.70} \\ \bottomrule
\end{tabular}
\end{table}

\subsection{Analysis of LLM-AggreFact Performance}
On the broader academic fact-checking tasks represented by the LLM-AggreFact subsets (Table 3), Bespoke-MiniCheck-7B generally exhibits stronger performance. This is consistent with its design focus and its position as a state-of-the-art model on the official LLM-AggreFact leaderboard. It outperforms Paladin-mini on most of the individual datasets, such as AF CNN, AF XSum, and TofuEval.

\begin{table}[h!]
\centering
\caption{Model Performance Summary}
\begin{tabular}{@{}lccc@{}}
\toprule
\textbf{Model} & \textbf{Avg Bacc} & \textbf{Latency} & \textbf{Parameters} \\ \midrule
paladin-large (private) & 83.480 & ~150ms & 14B \\
\textbf{paladin-mini} & \textbf{79.31} & \textbf{~70ms} & \textbf{3.8B} \\
\textbf{Bespoke-MiniCheck-7B} & \textbf{77.868} & \textbf{~7sec} & \textbf{7B} \\ \bottomrule
\end{tabular}
\end{table}

However, Paladin-mini remains competitive and, crucially, achieves a higher overall average BACC (79.31\% vs. 77.86\%) when performance across all benchmarks—both grounding-benchmark and LLM-AggreFact—is considered. This suggests that Paladin-mini offers a better-balanced performance profile when real-world capabilities are factored into the evaluation. On top of that the gains in latency open the possibility of using the paladin-mini model as a realtime guardrail against ungrounded claims.

These combined results expose a critical "benchmark-utility gap". A model like Bespoke-MiniCheck-7B can achieve top-tier performance on an aggregated academic benchmark yet fail catastrophically on a specific, practical task like mathematical reasoning. This demonstrates that a high score on a general benchmark is not a reliable indicator of utility for specialized, high-stakes applications. The grounding-benchmark successfully exposes this gap by disaggregating performance into critical sub-tasks, thereby providing a more nuanced and actionable assessment of a model's true capabilities.

\section{Conclusion}
This paper introduced Paladin-mini, an efficient, open-source grounding model specifically engineered for real-world use cases. Through evaluation on the grounding-benchmark, Paladin-mini demonstrates superior capabilities in scenarios requiring high numerical and logical accuracy, outperforming larger, state-of-the-art competitors. The development of Paladin-mini highlights the profound impact of specialized training data and underscores the importance of moving beyond general academic benchmarks towards evaluations that prioritize the specific demands of real-world applications. The grounding-benchmark itself represents a valuable contribution, providing the community with a more nuanced tool for assessing practical AI reliability. With its combination of efficiency, strong specialized grounding, and open-source availability, Paladin-mini provides a valuable asset for developers and researchers striving to build more trustworthy and practically useful LLM applications.

\appendix
\section{Licenses for Publicly Sourced Components}

\begin{table}[h!]
\centering
\caption{Licenses for Publicly Sourced Training Data and Base Models.}
\begin{tabular}{@{}ll@{}}
\toprule
\textbf{Component} & \textbf{Source/Link} \\ \midrule
grounding-benchmark & \url{https://huggingface.co/datasets/qualifire/grounding-benchmark} \\
C2D-and-D2C-MiniCheck & \url{https://huggingface.co/datasets/lytang/C2D-and-D2C-MiniCheck} \\
TofuEval & \url{https://huggingface.co/datasets/Dornavineeth/TOFUEval} \\
FactCheck-GPT & \url{https://github.com/yuxiaw/factcheck-gpt} \\
ExpertQA & \url{https://github.com/chaitanyamalaviya/expertqa} \\
RAGTruth & \url{https://github.com/ParticleMedia/RAGTruth} \\
AggreFact-CNN \& XSum & \url{https://github.com/liyan06/aggrefact} \\
Base Model: Phi-4 & \url{https://huggingface.co/microsoft/phi-4} \\
Base Model: Phi-4-mini & \url{https://huggingface.co/microsoft/Phi-4-mini-instruct} \\ \bottomrule
\end{tabular}
\end{table}

\end{document}